\documentclass[11pt]{article}
\usepackage[preprint]{acl}

\usepackage{times}
\usepackage{latexsym}
\usepackage{booktabs}
\usepackage{xcolor,colortbl}
\usepackage{graphicx}
\usepackage{multirow}
\usepackage{booktabs,threeparttable,xcolor,colortbl}
\usepackage{tcolorbox}
\tcbuselibrary{listings, breakable}

\definecolor{headerblue}{HTML}{B7C5E6}
\definecolor{datacellgreen}{HTML}{D9EAD3}
\definecolor{datacellpink}{HTML}{FFD1DC}

\usepackage[T1]{fontenc}

\usepackage[utf8]{inputenc}

\usepackage{microtype}

\usepackage{inconsolata}

\usepackage{graphicx}

\title{MetaBench: A Multi-task Benchmark for Assessing LLMs in Metabolomics}

\author{
 \textbf{Yuxing Lu\textsuperscript{1,2}},
 \textbf{Xukai Zhao\textsuperscript{3}},
 \textbf{J. Ben Tamo\textsuperscript{4}},
 \textbf{Micky C. Nnamdi\textsuperscript{4}},
\\
 \textbf{Rui Peng\textsuperscript{2}},
 \textbf{Shuang Zeng\textsuperscript{1,2}},
 \textbf{Xingyu Hu\textsuperscript{5}},
 \textbf{Jinzhuo Wang\textsuperscript{2,*}},
 \textbf{May D. Wang \textsuperscript{1,4,*}},
\\
\\
 \textsuperscript{1}Wallace H. Coulter Department of Biomedical Engineering, Georgia Institute of Technology and Emory University \\
 \textsuperscript{2}College of Future of Technology, Peking University \\
 \textsuperscript{3}School of Architecture, Tsinghua University \\
 \textsuperscript{4}School of Electrical and Computer Engineering, Georgia Institute of Technology \\
 \textsuperscript{5}School of Computer Science, Georgia Institute of Technology
\\
 \small{
   \textbf{Correspondence:} \href{wangjinzhuo@pku.edu.cn}{wangjinzhuo@pku.edu.cn}, \href{maywang@gatech.edu}{maywang@gatech.edu}
 }
}

\begin{document}
\maketitle
\begin{abstract}
Large Language Models (LLMs) have demonstrated remarkable capabilities on general text; however, their proficiency in specialized scientific domains that require deep, interconnected knowledge remains largely uncharacterized. Metabolomics presents unique challenges with its complex biochemical pathways, heterogeneous identifier systems, and fragmented databases. To systematically evaluate LLM capabilities in this domain, we introduce \textbf{MetaBench}, the first benchmark for metabolomics assessment. Curated from authoritative public resources, MetaBench evaluates five capabilities essential for metabolomics research: \emph{knowledge}, \emph{understanding}, \emph{grounding}, \emph{reasoning}, and \emph{research}. Our evaluation of 25 open- and closed-source LLMs reveals distinct performance patterns across metabolomics tasks: while models perform well on text generation tasks, cross-database identifier grounding remains challenging even with retrieval augmentation. Model performance also decreases on long-tail metabolites with sparse annotations. With MetaBench, we provide essential infrastructure for developing and evaluating metabolomics AI systems, enabling systematic progress toward reliable computational tools for metabolomics research.
\end{abstract}

\section{Introduction}

Large Language Models (LLMs) are being rapidly adopted across metabolomics research, driven by their demonstrated success in adjacent biomedical domains~\citep{liu2025application,bekbergenova2025metabot}. Biomedical LLMs have transformed protein structure prediction, clinically assist with diagnosis and treatment planning, and chemistry-focused systems support reaction prediction and molecular design~\citep{wang2023pre, zhao2023survey, hu2024rag}. Research groups now routinely utilize LLMs for tasks ranging from analyzing experiment results to generating study proposals. However, this rapid adoption has outpaced systematic evaluation: we lack a comprehensive understanding of which metabolomics tasks LLMs can reliably perform, where they fail, and why. This evaluation gap poses significant risks for a field where incorrect metabolite assignments or pathway interpretations can propagate through analysis pipelines and lead to false biological conclusions.

The consequences of deploying LLMs without proper evaluation extend beyond individual research errors. Metabolomics research demands capabilities that differ fundamentally from general text understanding and generation~\citep{bifarin2025large}. Researchers must integrate information across specialized databases such as the Human Metabolome Database (HMDB) and the Kyoto Encyclopedia of Genes and Genomes (KEGG) ~\citep{wishart2022hmdb, kanehisa2025kegg}, each employing distinct identifier systems and ontologies. Without knowing which of the capabilities current LLMs possess, researchers cannot make informed decisions about where to deploy these tools, what verification procedures to implement, or how to interpret AI-assisted results. 

Current biomedical evaluation frameworks cannot answer these questions~\citep{chen2025benchmarking,krithara2023bioasq,jin2019pubmedqa}. These benchmarks evaluate capabilities in natural language understanding but do not measure the specialized operations that metabolomics requires. For example, high performance on MedQA provides no evidence for reliability on identifier grounding, and failure on BioASQ does not preclude success on pathway description generation. Without evaluation frameworks designed for metabolomics-specific tasks, the field lacks criteria for model selection, failure mode identification, or systematic improvement.

In this paper, we present \textbf{MetaBench}, a comprehensive benchmark designed to systematically assess LLM capabilities across metabolomics tasks. MetaBench evaluates five capability levels essential for metabolomics research: \textit{Knowledge} (factual recall of metabolite properties), \textit{Understanding} (generation of coherent pathway descriptions), \textit{Grounding} (accurate identifier mapping across heterogeneous databases), \textit{Reasoning} (extraction of structured relationships from natural language), and \textit{Research} (synthesis of comprehensive study descriptions). Through evaluation of 25 state-of-the-art models on \textasciitilde 8,000 test cases derived from authoritative resources including HMDB, KEGG, PathBank, MetaKG, and MetaboLights~\citep{wishart2022hmdb, guijas2018metabolomics,lu2025knowledge,metabolights}, we provide the first systematic characterization of LLM performance in metabolomics, revealing which capabilities current models possess, where critical bottlenecks exist, and what architectural innovations are needed. This work makes the following contributions:
\begin{itemize}
    \vspace{-7pt}
    \item We introduce MetaBench, the first comprehensive benchmark for evaluating LLM capabilities in metabolomics, comprising \textasciitilde 8,000 test cases across five core capability levels.
    \vspace{-7pt}
    \item We conduct a systematic evaluation of 25 LLMs, revealing how these models perform across different metabolomics tasks.
    \vspace{-7pt}
    \item We identify and analyze critical bottlenecks in current LLMs for metabolomics applications, providing mechanistic insights into failure modes and pathways for improvement.
\end{itemize}

\section{Related Work}
\subsection{Benchmarks in Scientific Domains}
The rapid advancement of LLMs has spurred the development of numerous benchmarks to assess their capabilities in the scientific domain, largely concentrating on broad domains like biomedicine, urban planning, psychology, and chemistry~\citep{cai2024sciassess,ren2024valuebench,zhao2025urban}. Biomedical benchmarks have traditionally focused on tasks like question answering over literature and named entity recognition from clinical notes~\citep{krithara2023bioasq, jin2019pubmedqa,jiang2025benchmarking}. While invaluable, these benchmarks do not capture the domain knowledge, data structures, and specific tasks central to metabolomics, such as biochemical pathways, study proposals, and metabolite identifier groundings.

Similarly, in chemistry, benchmarks like MoleculeNet~\citep{wu2018moleculenet} have emerged to evaluate models on tasks such as molecule property prediction and name conversion. These benchmarks are tailored to the conventions of chemistry, focusing on chemical structure representations (e.g., SMILES) and reaction kinetics, which only partially overlap with the challenges in metabolomics~\citep{roessner2009metabolomics}. Metabolomics requires an understanding not just of individual molecules but of their roles within complex, dynamic biological systems~\citep{weckwerth2003metabolomics}. Our MetaBench is the first to bridge this gap, offering a suite of tasks designed specifically to test the deep, systems-level capabilities required for metabolomics.

\begin{figure*}
    \centering
    \includegraphics[width=\linewidth]{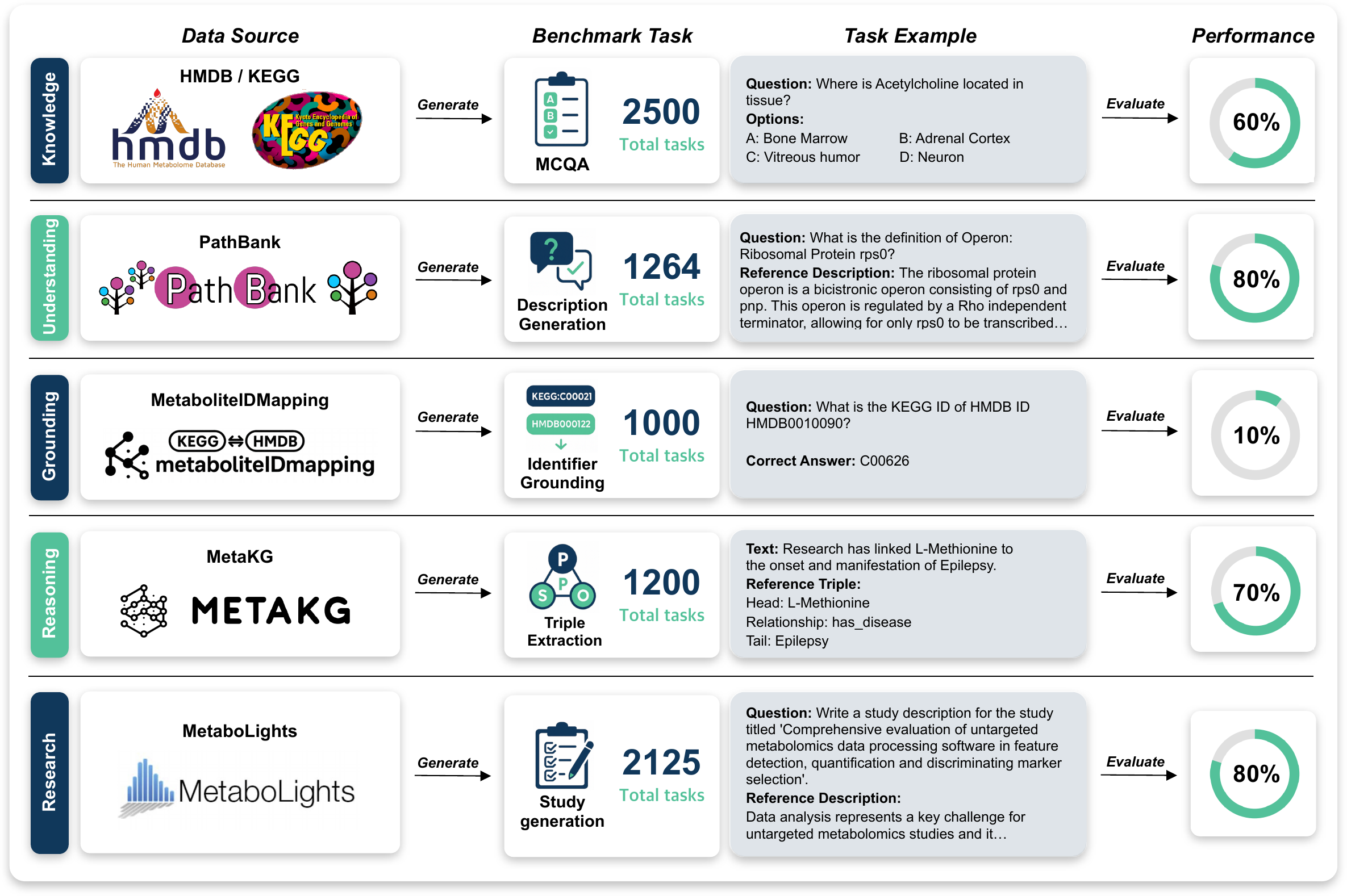}
    \caption{
    MetaBench construction and task design. MetaBench integrates data from multiple datasets to assess five capabilities: knowledge, understanding, grounding, reasoning, and research. 
    }
    \label{fig:1}
\end{figure*}

\subsection{NLP in Metabolomics}
The application of Natural Language Processing (NLP) in metabolomics is an emerging but critical area of research~\citep{bifarin2025large,coler2024metabolomics}. Prior work has primarily focused on leveraging NLP to aid in the interpretation of experimental results by contextualizing findings with existing literature or databases~\citep{lu2025knowledge,bekbergenova2025metabot,rahman2024incorporating}. These efforts often involve tools for named entity recognition (NER) to identify metabolites in text and relation extraction to link them to diseases or genes. However, these applications typically operate as pipeline components rather than end-to-end reasoning systems.

Furthermore, a significant challenge in the metabolomics field is knowledge fragmentation; information about a single metabolite may be spread across multiple databases (e.g., KEGG~\citep{kanehisa2025kegg}, HMDB~\citep{wishart2022hmdb}, PubChem~\citep{canese2013pubmed}), each using different identifier systems. This necessitates robust entity grounding to create a unified understanding~\citep{ji2020bert,weston2019named}. While entity grounding is a well-established NLP task, its application to the diverse and often overlapping identifiers in metabolomics is a unique and unsolved challenge. MetaBench formalizes these challenges into benchmark tasks for the first time, positioning our work as a necessary next step: moving from evaluating specific, isolated NLP tasks to a holistic assessment of powerful, end-to-end generative language models.

\section{MetaBench}

\subsection{Overview}
The lack of standardized evaluation benchmarks in metabolomics has left the field without clear criteria to assess LLM performance, making it difficult to compare methods, identify limitations, or provide reliable guidance for practical use. This gap slows both scientific progress and the safe adoption of LLMs in metabolomics research. To address this challenge, we curated a comprehensive benchmark from different sources (detailed introduction in \textbf{Appendix}~\ref{sec:knowledge-sources}) that evaluates LLMs across five capability levels, from factual knowledge recall to research-oriented text generation (\textbf{Figure}~\ref{fig:1}). This benchmark establishes the first critical foundation for methodological innovation and future applications of LLMs in metabolomics.

\subsection{Benchmark Taxonomy}
We organize MetaBench into five capabilities, each aligned with a specialized requirement in metabolomics workflows:

\paragraph{Knowledge.} Factual recall of general knowledge, such as the correct metabolite taxonomy.
\vspace{-7pt}
\paragraph{Understanding.} Generation of coherent descriptions of certain metabolites and pathways;
\vspace{-7pt}
\paragraph{Grounding.} Accurate alignment of identifiers across biomedical databases;
\vspace{-7pt}
\paragraph{Reasoning.} Entity and relationship extraction and structuring from natural language;
\vspace{-7pt}
\paragraph{Research.} Generation of research-related study descriptions from minimal prompts.

Together, these datasets provide a quantifiable, hierarchical framework for assessing LLMs in metabolomics, analogous to capability ladders used in general NLP benchmarks~\citep{jiang2025hibench}.

\begin{figure}
    \centering
    \includegraphics[width=\linewidth]{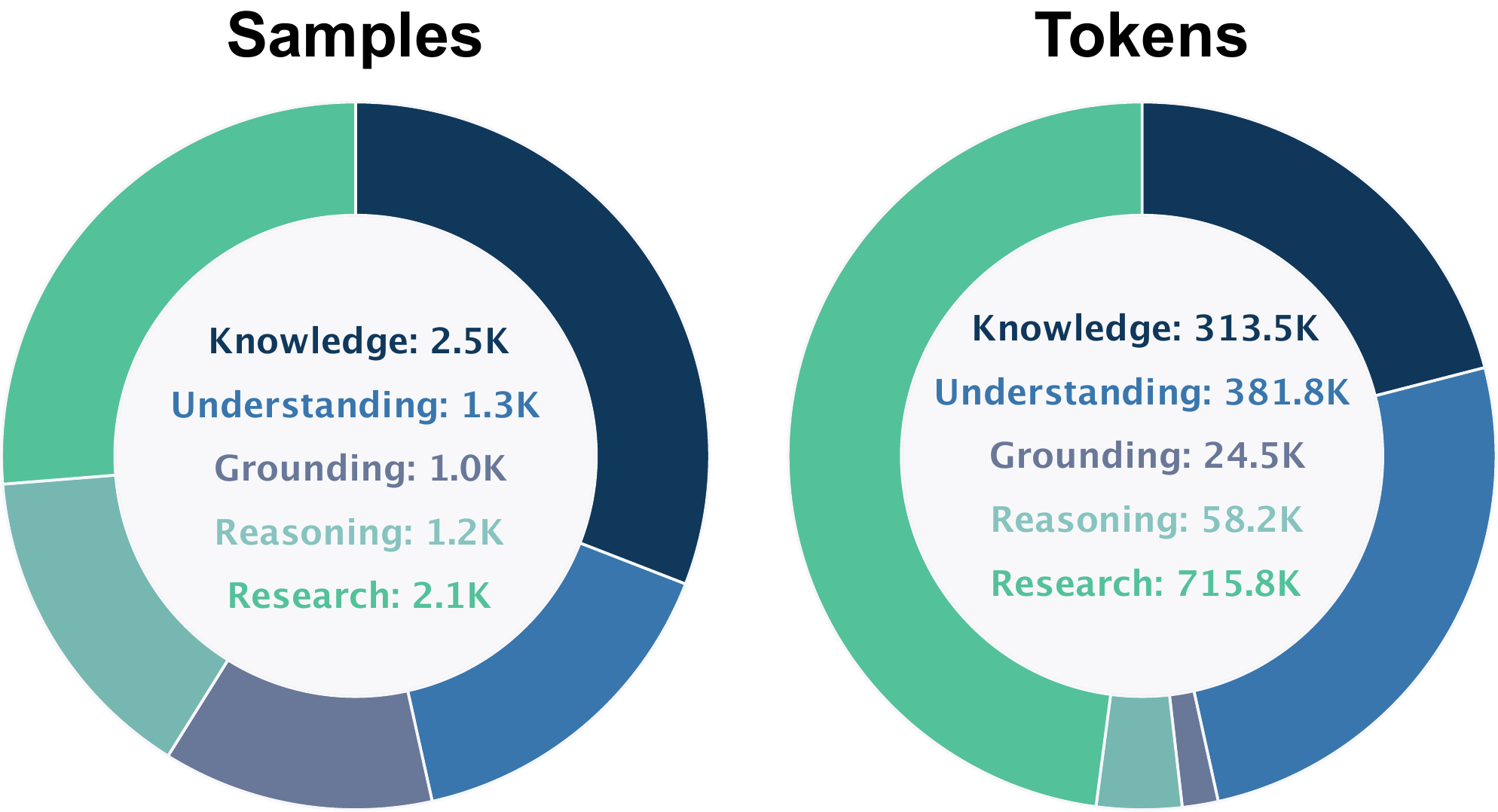}
    \caption{The statistics of MetaBench datasets.}
    \label{fig:dataset-stats}
\end{figure}

\subsection{Dataset Construction}
We instantiate these capabilities through public available datasets like HMDB~\citep{wishart2022hmdb}, KEGG~\citep{kanehisa2025kegg}, Metabolights~\citep{metabolights}, PathBank~\citep{pathbank}, and MetaKG~\citep{lu2025knowledge}, the overall process is illustrated in Figure \ref{fig:1}. All datasets are released publicly on HuggingFace\footnote{Anonymized for double-blind review} for reproducibility and the development of LLMs in metabolomics.

\subsubsection{Knowledge-based MCQA}
To assess how well different LLMs encode metabolomics knowledge, we construct a multiple-choice QA benchmark derived from HMDB~\citep{wishart2022hmdb} and KEGG~\citep{kanehisa2025kegg}. We select 26 attributes from these databases spanning taxonomy, molecular properties, biological associations, and pathway relationships. For each attribute, we extract entry-attribute pairs and generate four-option questions by combining the correct value with three in-domain distractors randomly sampled from other values of the same attribute. All questions follow standardized templates (\textbf{Appendix} \textbf{Table}~\ref{tab:appendix-MCQA-templates}), such as "What is the taxonomy class of {subject}?" The resulting dataset comprises 2,500 questions, with 100 questions per attribute. Examples are provided in \textbf{Appendix}~\ref{sec:appendix-mcqa-examples}.

\subsubsection{Description generation}
Beyond factual knowledge assessment, we evaluate whether LLMs can generate coherent, scientifically accurate descriptions of metabolomics concepts. This task tests the model's ability to produce informative pathway descriptions from pathway names alone. We curated a benchmark using 1,264 pathway-description pairs from PathBank~\citep{pathbank}, where each pathway name serves as the input prompt and the corresponding expert-written description serves as the reference output. Descriptions range from 148 to 6,452 tokens and contain comprehensive information about biochemical mechanisms, participating enzymes, and biological significance. Performance is assessed using BERTScore to measure semantic similarity between generated and reference texts. Examples are provided in \textbf{Appendix}~\ref{sec:appendix-qa-examples}.

\begin{table}[t]
\centering
\renewcommand{\arraystretch}{0.8}
\setlength{\tabcolsep}{1.4pt}
\caption{Statistics of the MetaBench datasets. Min / Avg. / Max indicates token lengths per sample.}
\label{tab:dataset-stats}
\begin{tabular}{l l c c c}
\toprule
\rowcolor{headerblue}
\textbf{Source} & \textbf{Capability} & \textbf{Min} & \textbf{Avg.} & \textbf{Max} \\
\midrule
HMDB, KEGG        & Knowledge        & 15 & 48.17  & 2113 \\
PathBank    & Understanding     & 26 & 166.28 & 880 \\
MetabolitesID  & Grounding       & 7  & 11.94  & 35 \\
MetaKG & Reasoning             & 9  & 17.95  & 35 \\
MetaboLights     & Research   & 17 & 222.57 & 852 \\
\bottomrule
\end{tabular}
\renewcommand{\arraystretch}{1.0} 
\end{table}
\subsubsection{Cross-Database Identifier Grounding}
A fundamental challenge in metabolomics research is integrating information about metabolites across heterogeneous databases, which requires accurate mapping between different identifier systems. Only through successful identifier resolution can fragmented knowledge be unified. To evaluate this grounding capability, we curated a benchmark using the MetabolitesID\footnote{https://github.com/yigbt/metaboliteIDmapping} package that requires models to map metabolite identifiers across KEGG, HMDB, and ChEBI databases. The task encompasses bidirectional mappings: KEGG$\leftrightarrow$HMDB, KEGG$\leftrightarrow$ChEBI, HMDB$\leftrightarrow$ChEBI, and identifier$\leftrightarrow$ name conversions. We sample 1,000 mapping pairs from a comprehensive table containing 130,005 entries, ensuring balanced representation across all mapping types. Each question provides a source identifier and requires the model to predict the corresponding target identifier, evaluated using exact match accuracy. Examples are provided in \textbf{Appendix}~\ref{sec:appendix-grounding-examples}.

\begin{table*}[t]
\centering
\begin{threeparttable}
\caption{MetaBench results for open- and closed-source LLMs across five capabilities. Cells report scores on a 0–100 scale. Best per column is shaded pink and second-best green.}
\label{tab:llm_eval_final}
\setlength{\tabcolsep}{3.1pt}
\renewcommand{\arraystretch}{0.73}
\setlength{\heavyrulewidth}{1pt}
\setlength{\lightrulewidth}{0.5pt}
\begin{tabular}{@{}lcccccc@{}}
\toprule
\rowcolor{headerblue}
\textbf{Model} & \textbf{Knowledge} & \textbf{Understanding} & \textbf{Grounding} & \textbf{Reasoning} & \textbf{Research} & \textbf{Average} \\
\midrule
\rowcolor{black!6} \textbf{Open-source models} & & & & & &\\
Qwen3-1.7b         & 32.01 & 81.46 & 0.27 & 56.08 & 79.46 & 49.86 \\
Qwen3-4b           & 35.94 & 83.21 & 0.27 & 66.39 & 80.52 & 53.27 \\
Qwen3-8b           & 43.58 & 82.87 & 0.25 & 66.47 & 80.83 & 54.40 \\
Qwen3-14b          & 52.62 & 82.95 & 0.29 & 65.14 & 80.88 & 56.38 \\
Qwen3-32b          & 56.42 & 83.01 & 0.33 & 67.72 & 82.57 & 58.01 \\
\midrule
Gemma-3-270m-it    & 13.61 & 81.28 & 0.20 & 11.17 & 80.37 & 37.33 \\
Gemma-3-1b-it      & 28.89 & 81.43 & 0.47 & 44.04 & 81.70 & 47.31 \\
Gemma-3-4b-it      & 46.26 & 81.97 & 0.53 & 68.50 & 82.29 & 55.91 \\
Gemma-3-12b        & 51.25 & 82.64 & 0.60 & 70.22 & 82.73 & 57.49 \\
Gemma-3-27b        & 55.82 & 83.05 & 0.70 & 72.69 & 83.24 & 59.10 \\
\midrule
Llama-3.2-1b       & 12.44 & 82.40 & 0.37 & 33.04 & 81.77 & 42.80 \\
Llama-3.2-3b       & 42.93 & 82.75 & 0.60 & 66.83 & 82.33 & 55.49 \\
Llama-3.1-8b       & 50.58 & 82.71 & 0.57 & 71.92 & 83.25 & 57.81 \\
Llama-3.1-70b      & 57.52 & 83.08 & \cellcolor{datacellgreen}0.63 & 72.98 & 83.18 & 59.88 \\
\midrule
DeepSeek-v3.1      & 54.34 & 82.38 & 0.48 & \cellcolor{datacellpink}73.81 & \cellcolor{datacellgreen}83.52 & 58.91 \\
\midrule
GPT-oss-20b & 50.66 & 82.83 & 0.50 & 70.16 & 82.87 & 57.40 \\
\midrule
\rowcolor{black!6} \textbf{Closed-source models} & & & & & &\\
GPT-4o-mini        & 53.82 & 82.36 & 0.27 & 66.39 & 80.02 & 56.17 \\
GPT-5-nano         & 35.65 & 82.71 & 0.27 & 68.47 & 82.00 & 53.82 \\
GPT-5-mini         & 59.02 & 82.95 & 0.40 & 71.89 & 82.39 & 59.33 \\
GPT-5              & \cellcolor{datacellgreen}60.50 & \cellcolor{datacellpink}83.52 & 0.47 & 72.39 & 82.72 & 59.92 \\
\midrule
Claude-haiku-3.5   & 44.34 & 82.15 & 0.47 & 70.64 & 81.41 & 55.80 \\
Claude-sonnet-3.7  & 59.78 & 82.87 & 0.53 & 71.31 & \cellcolor{datacellpink}83.87 & 59.67 \\
Claude-sonnet-4    & \cellcolor{datacellpink}60.94 & \cellcolor{datacellgreen}83.20 & \cellcolor{datacellpink}0.87 & 72.56 & 83.39 & \cellcolor{datacellpink}60.99 \\
\midrule
Gemini-2.5-flash   & 52.10 & 82.37 & 0.37 & 69.25 & 82.43 & 57.30 \\
Gemini-2.5-pro     & 59.54 & \cellcolor{datacellgreen}83.12 & \cellcolor{datacellgreen}0.63 & \cellcolor{datacellgreen}73.14 & 83.28 & \cellcolor{datacellgreen}60.34 \\
\bottomrule
\end{tabular}
\begin{tablenotes}[flushleft]
\footnotesize
\item Higher is better for all columns.
\end{tablenotes}
\end{threeparttable}
\end{table*}

\subsubsection{Knowledge extraction reasoning}
To evaluate structured knowledge extraction capabilities, we develop a benchmark that tests whether models can accurately parse metabolomics facts from natural language into knowledge graph triples. We select 1,200 triples from MetaKG~\citep{lu2025knowledge} spanning six relationship types: \texttt{has\_disease}, \texttt{has\_disposition}, \texttt{has\_smiles}, \texttt{has\_synonym}, \texttt{has\_class}, and \texttt{has\_tissue\_location}. For each triple, we use DeepSeek-V3.1~\citep{deepseekv3} to generate natural language sentences expressing the relationship. For example, the triple \textit{(Inosine, has\_tissue\_location, Platelet)} is rendered as "Within the human body, the blood's platelets are a known reservoir for the compound inosine." The generation prompt is provided in \textbf{Appendix}~\ref{sec:appendix-nl-generation}. Models must extract the complete (head, relationship, tail) triple from each sentence, requiring both natural language understanding and structured reasoning. Performance is evaluated using exact match accuracy, where all three components must be correctly identified. Representative examples are shown in \textbf{Appendix}~\ref{sec:appendix-reasoning-examples}.

\subsubsection{Study description generation }
The ultimate goal of applying LLMs to metabolomics is to enable comprehensive research support, from interpreting experimental results through multi-agent systems and data analysis pipelines to potentially designing or executing experiments autonomously. As a step toward this vision, we evaluate models on a research-level task requiring the generation of detailed study descriptions from concise titles. We curated a benchmark using 2,125 title-description pairs from MetaboLights~\citep{metabolights}, a leading metabolomics data repository. Each study title serves as the input prompt, and models must generate the corresponding comprehensive description containing experimental methodologies, analytical techniques, key findings, and biological implications. Examples are provided in \textbf{Appendix}~\ref{sec:appendix-research-examples}.

\subsection{Statistics}
\textbf{Table}~\ref{tab:dataset-stats} and \textbf{Figure}~\ref{fig:dataset-stats} present comprehensive statistics for the MetaBench datasets. We report minimum, average, and maximum token lengths per sample across all tasks. MetaBench comprises 8,100 samples distributed across five capability levels, with Knowledge (2,500 samples) and Research (2,125 samples) representing the largest subsets. Token volume varies substantially by task type: Understanding and Research tasks generate the highest token counts due to paragraph-level outputs (averaging 166 and 223 tokens, respectively), while Grounding and Reasoning tasks involve shorter structured responses (averaging 12 and 18 tokens). This distribution ensures comprehensive evaluation across diverse task formats, from concise factual retrieval and structured extraction to extended scientific text generation.

\subsection{Evaluation}
We employ task-appropriate metrics to ensure rigorous and meaningful evaluation. For classification tasks (knowledge MCQA, identifier grounding, and triple extraction reasoning), we report exact match accuracy. For generation tasks (pathway description generation and study description), we use BERTScore~\citep{zhang2019bertscore} (RoBERTa~\citep{liu2019roberta} as backbone model) to measure semantic similarity between generated and reference texts. All systems prompts for each task are provided in \textbf{Appendix}~\ref{app:system_prompts}.

For closed-source models, we perform inference through official APIs provided by OpenAI, Anthropic, and Google. For open-source models, we deploy them locally on H200 GPU clusters using the vLLM~\footnote{https://docs.vllm.ai/en/latest/} framework for efficient inference. We standardize all inference settings: temperature is set to 0.1 to encourage deterministic outputs, maximum generation length is capped at 4,096 tokens, and thinking modes are disabled for fair comparison. Each task uses a tailored system prompt that specifies the expected output format and evaluation criteria, ensuring alignment between model responses and metric requirements.

\section{Results}

\subsection{Overall performance across capabilities}
We evaluate 25 state-of-the-art LLMs spanning eight model families (\textbf{Appendix Table}~\ref{tab:models_metabench}). \textbf{Open-source models:} \textit{Qwen3~\citep{Qwen3}} (1.7B, 4B, 8B, 14B, 32B), \textit{Gemma-3~\citep{Gemma3}} (270M-it, 1B-it, 4B-it, 12B, 27B), \textit{Llama~\citep{Llama3}} (3.2-1B, 3.2-3B, 3.1-8B, 3.1-70B), \textit{DeepSeek-v3.1~\citep{deepseekv3}}, and \textit{GPT-oss-20b~\citep{GPToss}}. \textbf{Closed-source models:} \textit{GPT~\citep{GPT4,GPT4o}} (4o-mini, 5-nano, 5-mini, 5), \textit{Claude~\citep{Claude3,Claude4}} (haiku-3.5, sonnet-3.7, sonnet-4), and \textit{Gemini-2.5~\cite{Gemini}} (flash, pro). All evaluations use standard inference settings without external tools unless explicitly noted.

\textbf{Table}~\ref{tab:llm_eval_final} presents performance across five capabilities. Overall, \textit{Claude-sonnet-4} achieves the highest average score (60.99), followed closely by \textit{Gemini-2.5-pro} (60.34) and \textit{GPT-5} (59.92). Among open-source models, \textit{Llama-3.1-70b} leads at 59.88, nearly matching the best closed-source systems, followed by \textit{Gemma-3-27b} (59.10), \textit{DeepSeek-v3.1} (58.91), \textit{Qwen3-32b} (58.01), and \textit{GPT-oss-20b} (57.40). The narrow performance gap between top open-source and closed-source models (less than 1.5 points) demonstrates that open models can achieve competitive metabolomics reasoning when properly scaled.

\begin{figure}
    \centering
    \includegraphics[width=\linewidth]{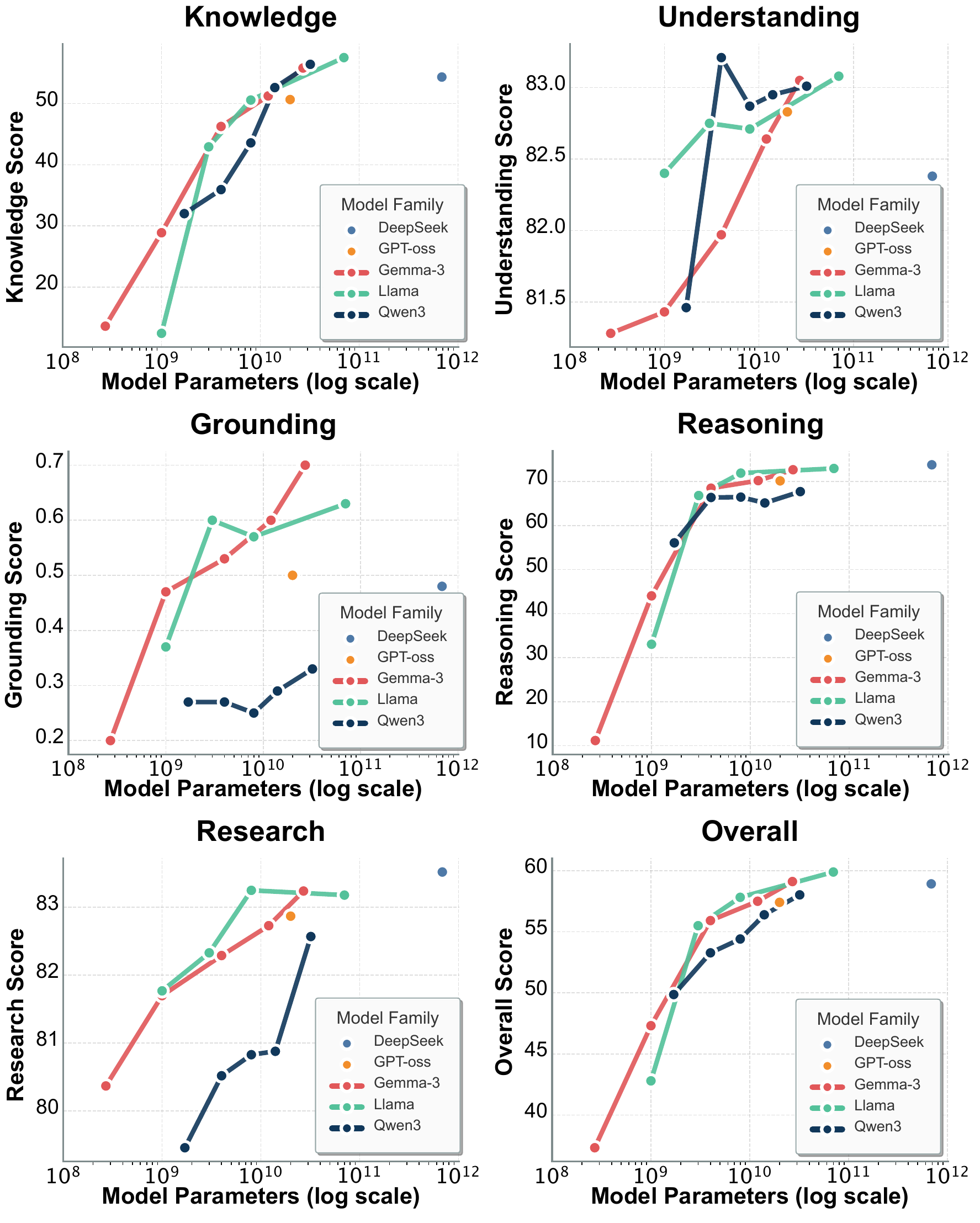}
    \caption{Parameter scaling trends in MetaBench. Average performance increases roughly log-linearly with model size, with consistent family-wise improvements.}
    \label{fig:3}
\end{figure}

Task-specific analysis reveals distinct capability profiles. For Knowledge tasks, \textit{Claude-sonnet-4} marginally leads at 60.94, slightly ahead of \textit{GPT-5} (60.50) and \textit{Gemini-2.5-pro} (59.54). Understanding shows remarkably compressed performance, with all competitive models clustering between 81–84\%; \textit{GPT-5} achieves the highest score (83.52), but the narrow range suggests this capability saturates early across modern architectures. Grounding presents a stark contrast: without retrieval augmentation, even the best model (\textit{Claude-sonnet-4}) achieves only 0.87\% accuracy, two orders of magnitude below other tasks. We analyze this bottleneck in detail in §\ref{sec:results-retrieval}. For Reasoning, \textit{DeepSeek-v3.1} excels at 73.81\%, with \textit{Gemini-2.5-pro} (73.14) and \textit{Claude-sonnet-4} (72.56) close behind. Finally, Research returns to a high, compressed band led by \textit{Claude-sonnet-3.7} (83.87), \textit{DeepSeek-v3.1} (83.52), and \textit{Claude-sonnet-4} (83.39).

\subsection{Parameter scaling trends}
The scaling law works on the five metabolomic capabilities, but the slope is task dependent (\textbf{Figure}~\ref{fig:3}, \textbf{Table}~\ref{tab:llm_eval_final}). Averaged across tasks, scores rise roughly with diminishing returns at the largest scales; family leaders at similar sizes differ by only 1–3 points, highlighting the influence of pretraining data and objectives beyond size alone. Family-wise scaling trends are consistent: \textit{Qwen3} improves from 49.86 (1.7B) to 58.01 (32B), \textit{Gemma-3} from 37.33 (270M-it) to 59.10 (27B), and \textit{Llama} from 42.80 (3.2-1B) to 59.88 (3.1-70B). \textit{GPT-20b-oss} and \textit{DeepSeek-v3.1} also demonstrate performance consistent with their parameter counts.

Task-level analysis shows distinct behaviors. Knowledge scales cleanly with size, with Llama rising from 12.44 to 57.52 and Qwen3 from 32.01 to 56.42. Reasoning also benefits strongly, where mid-size models plateau in the high 60s and the largest variants approach 73–74. By contrast, Understanding and Research tasks saturate early: most models fall in a narrow 81–84 band, with larger variants adding less than 2 points, suggesting these generation tasks depend more on instruction tuning than sheer scale. Finally, Grounding remains flat; its accuracy does not exceed 0.87\% even for the largest models, showing that parameter growth alone cannot resolve identifier mapping. These results underline that while scale reliably improves recall and reasoning, it delivers marginal gains on fluent generation already near ceiling and fails on grounding tasks unless combined with retrieval and schema-aware normalization.

\begin{figure}[t]
    \centering
    \includegraphics[width=\linewidth]{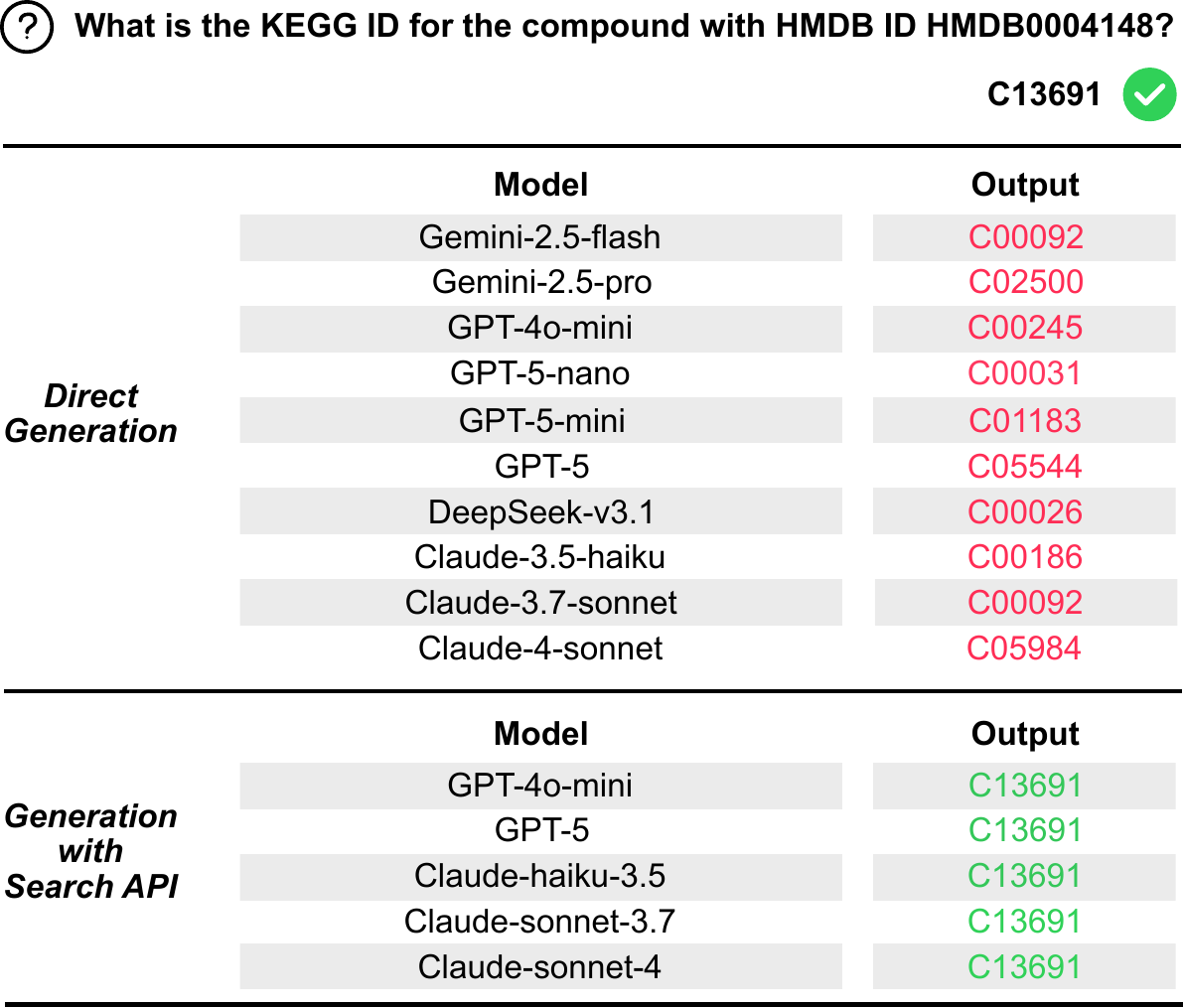}
    \caption{LLMs consistently fail in metabolite identifier grounding. 11 LLMs were asked to retrieve KEGG ID for HMDB0004148. None produced the correct answer. In contrast, when augmented with web search API, all successfully provided the correct mapping.}
    \vspace{-5pt}
    \label{fig:4}
\end{figure}

\subsection{Identifier grounding is the bottleneck}
\label{sec:results-retrieval}
Among all tested capabilities, Grounding is the lowest-performing capability. Without retrieval, accuracy remains near zero and does not exceed 0.87\% even for the largest models (\textbf{Table}~\ref{tab:llm_eval_final}). A probe illustrates the failure: asked “What is the KEGG ID for HMDB0004148?”, 11 LLMs answered incorrectly (e.g., \texttt{C00092}, \texttt{C02500}); none returned the correct \texttt{C13691} (\textbf{Figure}~\ref{fig:4}). With web search API, all five tested systems produced \texttt{C13691} on this item (\textbf{Figure}~\ref{fig:4}).

\begin{figure}[ht]
    \centering
    \includegraphics[width=\linewidth]{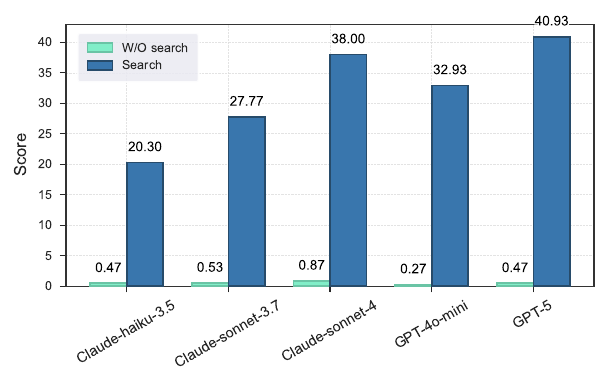}
    \caption{Without search capabilities (blue bars), all models achieve less than 1\% accuracy on cross-database metabolite mapping. Enabling web search (orange bars) improves accuracy by 40–150×, though performance remains below 41\% even for the best model (GPT-5), indicating that retrieval alone is insufficient to resolve the grounding bottleneck.}
    \label{fig:5}
\end{figure}

Aggregate results show the same pattern. Across five representative models, accuracy rises from \{0.27, 0.47, 0.53, 0.87\}\% without search to \{20.30, 27.77, 32.93, 38.00, 40.93\}\% with search (\textbf{Figure}~\ref{fig:5}). This 40–150× gain remains far from ceiling, indicating missing external evidence and normalization rather than decoding issues.

Grounding failures may be because the task combines sparse signals, lossy tokenization, and a misaligned objective under moving targets and ambiguous nomenclature. Metabolite IDs are rare in pretraining corpora and often confined to supplementary tables, so models learn weak parametric associations. Subword tokenizers fracture strings such as \texttt{HMDB0004148} into pieces (e.g., \texttt{HMD}, \texttt{B000}, \texttt{4148}), undermining exact matching. Next-token training optimizes plausibility rather than exact resolution, yielding confident but incorrect IDs. Concurrent database updates in HMDB and KEGG further skew memorized mappings. Added to this, synonyms, tautomers, salts, stereochemistry, and organism or compartment context create many-to-many name–ID relations that text-only models cannot reliably disambiguate. Retrieval with schema-aware normalization targets these causes; scaling alone does not.

\subsection{Long tail problem in metabolomics}
Metabolite databases such as HMDB exhibit information concentration: metabolites with lower HMDB IDs (typically discovered earlier) contain substantially more annotated attributes, while those with higher IDs show progressively sparser information~\citep{xia2022statistical}. This pattern reflects the field's trajectory from well-studied central metabolites toward incompletely characterized peripheral compounds.
\begin{figure}
    \centering
    \includegraphics[width=1\linewidth]{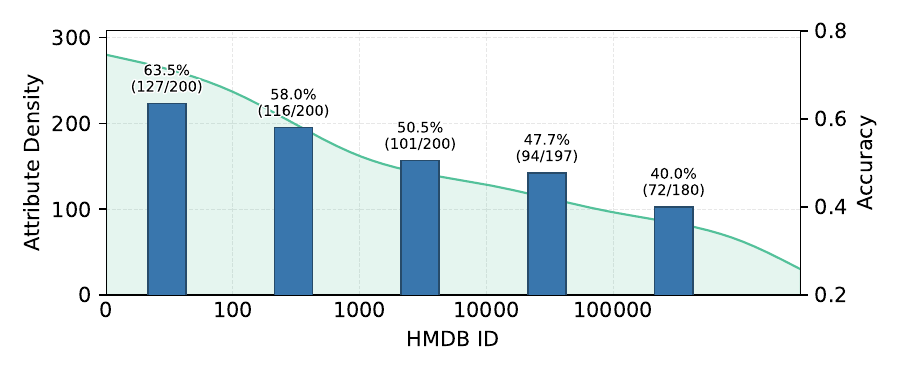}
    \caption{Long-tail distribution in metabolite knowledge. Attribute density (blue line) and model accuracy (green bars) both decline across HMDB ID ranges. }
    \label{fig:6}
    
\end{figure}

To quantify how this heterogeneity affects model performance, we stratified metabolites by HMDB ID ranges and measured average attribute density: <100 (236.73 attributes/metabolite), 100-1,000 (161.92), 1,000-10,000 (128.53), 10,000-100,000 (96.30), and >100,000 (63.38). We sampled 200 MCQA questions per bin and evaluated multiple LLMs. Results from \textit{GPT-oss-20b} are shown in \textbf{Figure}~\ref{fig:6}, where accuracy declines monotonically with decreasing density: 63.5\% (127/200) for IDs <100, 58.0\% (116/200) for IDs between 100 and 1000, 50.5\% for IDs between 1000 and 10000, 47.7\% for IDs between 10000 and 100000, versus 40.0\% (72/180) for IDs >100,000, resulting in a 23.5 percentage gap.

This long-tail effect reveals a fundamental challenge: models perform acceptably on well-studied metabolites dominating training corpora but fail disproportionately on the sparsely annotated compounds that constitute the metabolome's majority. Simply scaling training data cannot address the problem since the long tail reflects actual knowledge gaps in the field. Addressing this limitation may require active learning strategies prioritizing difficult cases, multi-modal models leveraging structural information when annotations are sparse, or uncertainty-aware systems that flag low-confidence predictions for long-tail metabolites~\citep{wang2024llm,kandpal2023large}.

\section{Discussion}

Our evaluation of 25 LLMs reveals a striking capability heterogeneity that challenges assumptions about LLM readiness for metabolomics. These results indicate that current training paradigms, optimized for fluent text generation, produce models that excel at synthesizing coherent scientific narratives while struggling with precise factual retrieval and structured knowledge operations~\citep{kalai2025language}. Most critically, the catastrophic failure on identifier Grounding (<1\% without retrieval, 41\% maximum with search API) represents a fundamental limitation rather than a data scarcity problem (more emphasis on the grounding tasks can refer to \textbf{Appendix}~\ref{sec:appendix-grounding-importance}). The grounding bottleneck has immediate practical implications: metabolomics applications requiring cross-database integration cannot rely on LLM capabilities alone and must implement specialized identifier resolution systems with schema-aware normalization and chemical structure reasoning.

Beyond diagnosing current limitations, MetaBench establishes a framework for targeted model improvement and responsible deployment in metabolomics. Importantly, our benchmark dataset construction method provides a replicable pathway for creating domain-specific fine-tuning (DFT) datasets beyond evaluation. As new models emerge, whether general LLMs with improved scientific reasoning or domain-specific models pretrained on metabolomics corpora, MetaBench enables continuous, standardized assessment beyond aggregate scores that can obscure critical limitations. By providing both evaluation infrastructure and a concrete methodology for dataset construction, MetaBench supports the development of more capable and reliable AI systems for metabolomics research.

\section{Conclusion}
We present MetaBench, the first comprehensive benchmark for evaluating LLMs on metabolomics. Through the evaluation of 25 LLMs across five capabilities, we reveal substantial performance variations across models and demonstrate that while model scaling improves reasoning, current models catastrophically fail on cross-database mapping and long-tail generalization, establishing that metabolomics requires precision and structured knowledge integration beyond current architectures and training corpus. By publicly releasing MetaBench, we provide essential infrastructure for developing scientifically-grounded models capable of supporting real-world metabolomics research and enabling systematic progress toward reliable AI-assisted discovery.

\section{Limitation}
Since there is currently no publicly available metabolomics-specialized LLM; consequently, MetaBench evaluates general-purpose and broad biomedical models, quantifying cross-domain generalization and establishing a strong baseline for future in-domain systems. The benchmark centers on widely used resources (KEGG, HMDB, ChEBI, PathBank, MetaboLights), choices that favor clean, reproducible comparisons while deferring spectra/structure modalities and additional databases to later releases. We score classification with accuracy and generation with BERTScore to enable scale and consistency. Results are produced under a standardized, tool-free decoding setup, with retrieval analyzed separately, to isolate intrinsic model behavior. These constraints are deliberate and highlight the release’s strengths: clarity, reproducibility, and extensibility, while charting a direct path to add specialized models in future versions.

\section{Potential risks}
While MetaBench establishes evaluation infrastructure for metabolomics AI, we acknowledge considerations for responsible use. Our transparent reporting of performance disparities, such as the identifier grounding challenge and 23.5\% variation across metabolite coverage, helps prevent premature deployment while guiding targeted improvements. The modular design enables continuous benchmark evolution as models advance, and our curation from established resources ensures tasks reflect real metabolomics workflows. By providing granular evaluation across five capabilities rather than single scores, MetaBench enables informed decisions about where LLMs can augment research and where human expertise remains essential, establishing not just current baselines but the measurement framework necessary for systematic advancement of AI in metabolomics.

\bibliography{custom}

\clearpage

\appendix

\section{Sources for MetaBench}
\label{sec:knowledge-sources}
MetaBench integrates data from six specialized metabolomics resources, each serving distinct roles in the field. This section provides detailed descriptions of these sources and their contributions to our benchmark.
\subsection{Human Metabolome Database (HMDB)}
The Human Metabolome Database (HMDB)~\citep{wishart2022hmdb} is the most comprehensive metabolomics database focused on human metabolism, cataloging over 220,000 metabolite entries including both endogenous metabolites and exogenous compounds. Each entry provides extensive biochemical information including molecular descriptors (SMILES, InChI), physicochemical properties, tissue and biofluid locations, disease associations, and literature references using standardized identifiers (\texttt{HMDB\#\#\#\#\#\#\#}). In MetaBench, HMDB serves as a primary source for the Knowledge MCQA task, providing ground truth data for 26 distinct metabolite attributes, and forms one node of the cross-database identifier mapping in the Grounding capability assessment.

\subsection{Kyoto Encyclopedia of Genes and Genomes (KEGG)}
The Kyoto Encyclopedia of Genes and Genomes (KEGG)~\citep{kanehisa2025kegg} integrates genomic, chemical, and systemic functional information, with KEGG COMPOUND containing over 18,000 metabolite entries organized within the context of biochemical pathways, reactions, and molecular interactions. Each compound receives a KEGG identifier (\texttt{C\#\#\#\#\#}) and is annotated with pathway memberships, enzyme associations (EC numbers), reaction participation, and orthology relationships. 

\subsection{PathBank}
PathBank~\citep{pathbank} is a comprehensive visual database containing over 110,000 detailed pathway diagrams and expert-curated descriptions covering metabolic, signaling, drug action, and disease pathways across multiple organisms, with particular emphasis on human metabolism. Each entry contains a pathway name, comprehensive textual description ranging from concise summaries to extensive multi-paragraph explanations (148–6,452 tokens), interactive diagrams, and lists of participating metabolites and proteins. PathBank serves as the exclusive source for the Understanding capability assessment in MetaBench, where 1,264 pathway name-description pairs evaluate whether models can generate scientifically accurate, coherent descriptions of biochemical pathways, testing conceptual understanding beyond simple factual recall.

\subsection{MetabolitesIDMapping}
MetaboliteIDmapping is a specialized mapping resource providing comprehensive cross-references between metabolite identifiers across major databases, addressing the fundamental challenge of identifier heterogeneity in metabolomics, where the same metabolite may be referenced differently across resources. The mapping table contains over 130,000 verified equivalence relationships linking identifiers from KEGG, HMDB, ChEBI, PubChem, CAS Registry Numbers, and chemical names (IUPAC and common), enabling bidirectional translation between identifier systems. MetabolitesID forms the foundation of the Grounding capability assessment, where 1,000 sampled identifier pairs construct mapping tasks that directly evaluate a critical bottleneck in practical metabolomics workflows: integrating fragmented knowledge across heterogeneous resources.

\subsection{MetaKG}
MetaKG~\citep{lu2025knowledge} is a large-scale metabolomics knowledge graph structuring information from multiple databases into a unified semantic network with over 2 million entities and 10 million relationships, representing the most comprehensive structured representation of metabolomics knowledge to date. The heterogeneous graph contains nodes representing metabolites, diseases, proteins, genes, tissues, and pathways, connected by typed relationships, with each triple derived from authoritative sources and validated for consistency. MetaKG serves as the source for the Reasoning capability assessment, where 1,200 triples spanning six relationship types are converted into natural language sentences, requiring models to parse unstructured text and extract structured (head, relationship, tail) triples, testing both natural language understanding and structured reasoning essential for knowledge extraction from scientific literature.

\subsection{MetaboLights}
MetaboLights~\citep{metabolights} is the largest open-access metabolomics data repository hosted by the European Bioinformatics Institute (EMBI-EBI), containing over 7,000 studies with associated metadata, analytical methods, and raw data files that serve as a comprehensive archive for metabolomics experimental data. Each study (\texttt{MTBLS\#\#\#\#}) includes structured submissions with study titles, detailed descriptions, experimental designs, sample information, analytical protocols (LC-MS, GC-MS, NMR), data processing workflows, and expert-written descriptions providing comprehensive information about research objectives, methodologies, key findings, and biological significance. MetaboLights provides the foundation for the Research capability assessment, where 2,125 study title-description pairs evaluate whether models can generate comprehensive, research-level documentation from minimal prompts, testing capabilities essential for advanced metabolomics research support including experimental design synthesis, analytical technique knowledge, and scientific interpretation.

\section{MCQA templates}
\label{sec:appendix}

To construct the Knowledge Graph QA dataset, we selected 26 different relations from MetaKG, the curated metabolomics knowledge graph. For each relation, we designed a specific question template in natural language. These templates convert structured triples \textit{(head, relation, tail)} into interrogative forms that can be used for multiple-choice question answering.

\textbf{Table}~\ref{tab:appendix-MCQA-templates} lists the relation types and their associated templates. For example, the relation \texttt{has\_class} is expressed as ``What is the taxonomy class of \{subject\}?'' and \texttt{has\_disease} as ``Which disease is associated with \{subject\}?''. This template-based construction ensures consistency across questions while covering diverse relation types in metabolomics.

\begin{table*}[ht]
\centering
\begin{tabular}{|p{5.5cm}|p{9.5cm}|}
\hline
\textbf{Attribute} & \textbf{Question Template} \\
\hline
has\_class & What is the taxonomy class of \{subject\}? \\
has\_name & What is the name of \{subject\}? \\
has\_disease & Which disease is associated with \{subject\}? \\
has\_substituent & What is a substituent of \{subject\}? \\
has\_inchi & What is the InChI of \{subject\}? \\
has\_disposition & What is the disposition of \{subject\}? \\
has\_uniprot\_id & What is the UniProt ID of \{subject\}? \\
has\_synonym & What is a synonym for \{subject\}? \\
has\_process & What process is \{subject\} involved in? \\
has\_enzyme & Which enzyme is associated with \{subject\}? \\
related\_to\_protein & Which protein is \{subject\} related to? \\
has\_smiles & What is the SMILES notation of \{subject\}? \\
has\_cellular\_location & Where is \{subject\} located in the cell? \\
average\_molecular\_weight & What is the average molecular weight of \{subject\}? \\
chemical\_formula & What is the chemical formula of \{subject\}? \\
related\_to\_pathway & Which KEGG pathway is \{subject\} related to? \\
has\_inchikey & What is the InChIKey of \{subject\}? \\
has\_kegg\_id & What is the KEGG ID of \{subject\}? \\
has\_biospecimen\_location & Where is \{subject\} found as a biospecimen? \\
has\_pathway & What is a KEGG pathway associated with \{subject\}? \\
belongs\_to\_orthology & To which KEGG orthology does \{subject\} belong? \\
has\_reaction & What is a KEGG reaction involving \{subject\}? \\
has\_description & What is the description of \{subject\}? \\
belongs\_to\_network & To which KEGG network does Gene \{subject\} belong? \\
has\_tissue\_location & Where is \{subject\} located in tissue? \\
is\_a\_sub\_class\_of & \{subject\} is a subclass of what? \\
\hline
\end{tabular}
\caption{Relation types and their corresponding question templates.}
\label{tab:appendix-MCQA-templates}
\end{table*}

\section{Knowledge MCQA examples}
\label{sec:appendix-mcqa-examples}
The Knowledge capability is assessed through multiple-choice question answering (MCQA) tasks derived from HMDB and KEGG. Below are representative examples across different attribute types, illustrating the diversity of metabolomics knowledge tested in MetaBench. Questions span metabolite taxonomy, molecular properties, biological associations, pathway relationships, and tissue locations. Each question provides four options with one correct answer, where distractors are sampled from valid values of the same attribute to ensure domain relevance and task difficulty.
\begin{tcolorbox}[
title=\textbf{Example 1: Tissue Location},
colframe=black!50,
colback=black!5,
boxrule=0.5pt,
arc=2pt,
fonttitle=\bfseries,
breakable
]
\textbf{Question:} Where is Acetylcholine located in tissue?

\textbf{Options:}
\begin{itemize}
\item A: Bone Marrow
\item B: Adrenal Cortex
\item C: Vitreous humor
\item D: Neuron
\end{itemize}
\textbf{Answer:} D
\end{tcolorbox}
\begin{tcolorbox}[
title=\textbf{Example 2: Taxonomy Classification},
colframe=black!50,
colback=black!5,
boxrule=0.5pt,
arc=2pt,
fonttitle=\bfseries,
breakable
]
\textbf{Question:} What is the taxonomy class of 2-Hydroxyestrone?

\textbf{Options:}
\begin{itemize}
\item A: Aralkylamines
\item B: Estrane steroids
\item C: Catechols
\item D: Very long-chain fatty acids
\end{itemize}
\textbf{Answer:} B
\end{tcolorbox}
\begin{tcolorbox}[
title=\textbf{Example 3: Database Identifier},
colframe=black!50,
colback=black!5,
boxrule=0.5pt,
arc=2pt,
fonttitle=\bfseries,
breakable
]
\textbf{Question:} What is the KEGG ID of Tetrahydrobiopterin?

\textbf{Options:}
\begin{itemize}
\item A: C00864
\item B: C00234
\item C: C01747
\item D: C00272
\end{itemize}
\textbf{Answer:} D
\end{tcolorbox}
\begin{tcolorbox}[
title=\textbf{Example 4: Enzyme Association},
colframe=black!50,
colback=black!5,
boxrule=0.5pt,
arc=2pt,
fonttitle=\bfseries,
breakable
]
\textbf{Question:} Which enzyme is associated with Glycine?

\textbf{Options:}
\begin{itemize}
\item A: 2.6.1.94
\item B: 2.1.1.156
\item C: 1.1.1.228
\item D: 4.2.3.87
\end{itemize}
\textbf{Answer:} B
\end{tcolorbox}
\begin{tcolorbox}[
title=\textbf{Example 5: Pathway Relationship},
colframe=black!50,
colback=black!5,
boxrule=0.5pt,
arc=2pt,
fonttitle=\bfseries,
breakable
]
\textbf{Question:} What is a pathway associated with C15976?

\textbf{Options:}
\begin{itemize}
\item A: SMP00497
\item B: SMP00417
\item C: SMP00384
\item D: SMP00507
\end{itemize}
\textbf{Answer:} C
\end{tcolorbox}

\section{Description generation examples}
\label{sec:appendix-qa-examples}
The Understanding capability is assessed through pathway description generation tasks derived from PathBank. Models receive pathway names as input and must generate comprehensive, scientifically accurate descriptions explaining the biochemical mechanisms, participating molecules, biological significance, and clinical relevance. Below are representative examples demonstrating the range of pathway types and description complexity in MetaBench, from metabolic pathways and disease mechanisms to drug action pathways and gene regulation systems.
\begin{tcolorbox}[
title=\textbf{Example 1: Signaling Pathway},
colframe=black!50,
colback=black!5,
boxrule=0.5pt,
arc=2pt,
fonttitle=\bfseries,
breakable
]
\textbf{Question:} What is the definition of Succinate Signalling?

\textbf{Reference Description:}
Tricarboxylic acid (TCA) cycle intermediates can function as inflammatory signals. Succinate enhances glycolysis in several ways. It inhibits prolyl hydroxylase domain (PHD) enzyme function, both directly by product inhibition and indirectly via reactive oxygen species (ROS), driving hypoxia-inducible factor-1$\alpha$ (HIF-1$\alpha$) accumulation and increased glycolysis. ROS also inhibit mitochondrial function, boosting glycolysis as a result. Elevated HIF-1$\alpha$ enhances the expression of genes containing HIF response elements (HREs), including the interleukin 1$\beta$ (IL-1$\beta$) gene. Succinate can signal through succinate receptor 1 (SUCNR1) and act in synergy with Toll-like receptors (TLRs) to boost dendritic cell function. NAD+ exerts several anti-inflammatory effects by activating sirtuins, a class of NAD+-dependent deacetylases. Finally, inactivation of Sirt3 has been demonstrated to enhance NLR family, pyrin domain containing 3 (NLRP3) inflammasome activation.
\end{tcolorbox}
\begin{tcolorbox}[
title=\textbf{Example 2: Gene Regulation},
colframe=black!50,
colback=black!5,
boxrule=0.5pt,
arc=2pt,
fonttitle=\bfseries,
breakable
]
\textbf{Question:} What is the definition of Operon: Ribosomal Protein rps0?

\textbf{Reference Description:}
The ribosomal protein operon is a bicistronic operon consisting of rps0 and pnp. This operon is regulated by a Rho independent terminator, allowing for only rps0 to be transcribed if the terminator is formed.
\end{tcolorbox}
\begin{tcolorbox}[
title=\textbf{Example 3: Disease Pathway},
colframe=black!50,
colback=black!5,
boxrule=0.5pt,
arc=2pt,
fonttitle=\bfseries,
breakable
]
\textbf{Question:} What is the definition of Glycogenosis, Type IA. Von Gierke Disease?

\textbf{Reference Description:}
Glycogen storage disease type I (also known as GSDI or von Gierke disease) is an inherited disorder caused by the buildup of a complex sugar called glycogen in the body's cells. The accumulation of glycogen in certain organs and tissues, especially the liver, kidneys, and small intestines, impairs their ability to function normally. Researchers have described two types of GSDI, which differ in their signs and symptoms and genetic cause. These types are known as glycogen storage disease type Ia (GSDIa) and glycogen storage disease type Ib (GSDIb). Two other forms of GSDI have been described, and they were originally named types Ic and Id. However, these types are now known to be variations of GSDIb; for this reason, GSDIb is sometimes called GSD type I non-a. Mutations in two genes, G6PC and SLC37A4, cause GSDI.
\end{tcolorbox}
\begin{tcolorbox}[
title=\textbf{Example 4: Drug Action Pathway},
colframe=black!50,
colback=black!5,
boxrule=0.5pt,
arc=2pt,
fonttitle=\bfseries,
breakable
]
\textbf{Question:} What is the definition of Tetracycline Action Pathway?

\textbf{Reference Description:}
Tetracycline is a short-acting antibiotic that is semi-synthetically produced from chlortetracycline, a compound derived from Streptomyces aureofaciens. Tetracycline enters bacterial cells by passively diffusing through membrane porin channels. Once inside the cell, tetracycline reversibly binds to the 30S subunit just above the binding site for aminoacyl tRNA. At its primary binding site, interactions with the sugar phosphate backbone of residues in helices 31 and 34 via hydrogen bonds with oxygen atoms and hydroxyl groups on the hydrophilic side of the tetracycline help anchor the drug in position. Salt bridge interactions between the backbone of 16S rRNA and tetracycline are mediated by a magnesium ion in the binding site. Tetracycline prevents incoming aminoacyl tRNA from binding to the A site on the ribosome-RNA complex via steric hindrance. This causes inhibition of protein synthesis and hence bacterial cell growth.
\end{tcolorbox}
\begin{tcolorbox}[
title=\textbf{Example 5: Pharmaceutical Mechanism},
colframe=black!50,
colback=black!5,
boxrule=0.5pt,
arc=2pt,
fonttitle=\bfseries,
breakable
]
\textbf{Question:} What is the definition of Sufentanil Action Pathway?

\textbf{Reference Description:}
Sufentanil is a pharmacologically-active synthetic small molecule derived from fentanyl and belongs to a class of drugs called opioids. Opioids are therapeutically employed to achieve analgesia. Sufentanil's rapid mechanism of action primarily involves its agonistic effects on mu-type opioid receptors which are inhibitory G-coupled protein receptors and lead to the inhibition of adenylate cyclase and decrease in cAMP production. It also inhibits nociceptive neurotransmitter release and induces membrane hyperpolarization. Analgesia, anesthesia, and respiratory depression are a consequence of remifentanial's action.
\end{tcolorbox}

\section{Grounding (Id mapping) examples}
\label{sec:appendix-grounding-examples}
The Grounding capability is assessed through cross-database identifier mapping tasks derived from MetabolitesID. This task evaluates whether models can accurately translate metabolite identifiers across heterogeneous database systems (KEGG, HMDB, ChEBI) and between structured identifiers and chemical names. Successful identifier grounding is fundamental to integrating fragmented metabolomics knowledge, yet represents a critical bottleneck for current LLMs. Below are representative examples demonstrating the diverse mapping types required in MetaBench.
\begin{tcolorbox}[
title=\textbf{Example 1: HMDB to KEGG Mapping},
colframe=black!50,
colback=black!5,
boxrule=0.5pt,
arc=2pt,
fonttitle=\bfseries,
breakable
]
\textbf{Question:} What is the KEGG ID of HMDB ID HMDB0010090?

\textbf{Answer:} C00626
\end{tcolorbox}
\begin{tcolorbox}[
title=\textbf{Example 2: KEGG to HMDB Mapping},
colframe=black!50,
colback=black!5,
boxrule=0.5pt,
arc=2pt,
fonttitle=\bfseries,
breakable
]
\textbf{Question:} What is the HMDB ID of KEGG ID C07251?

\textbf{Answer:} HMDB0014982
\end{tcolorbox}
\begin{tcolorbox}[
title=\textbf{Example 3: Name to KEGG Mapping},
colframe=black!50,
colback=black!5,
boxrule=0.5pt,
arc=2pt,
fonttitle=\bfseries,
breakable
]
\textbf{Question:} What is the KEGG ID of metabolite Ponalactone A?

\textbf{Answer:} C09174
\end{tcolorbox}
\begin{tcolorbox}[
title=\textbf{Example 4: Name to HMDB Mapping},
colframe=black!50,
colback=black!5,
boxrule=0.5pt,
arc=2pt,
fonttitle=\bfseries,
breakable
]
\textbf{Question:} What is the HMDB ID of metabolite Ethylparaben?

\textbf{Answer:} HMDB0032573
\end{tcolorbox}
\begin{tcolorbox}[
title=\textbf{Example 5: Name to ChEBI Mapping},
colframe=black!50,
colback=black!5,
boxrule=0.5pt,
arc=2pt,
fonttitle=\bfseries,
breakable
]
\textbf{Question:} What is the ChEBI ID of metabolite Croconazole?

\textbf{Answer:} 3920
\end{tcolorbox}
\vspace{0.5cm}
\noindent\textbf{Note:} These examples illustrate the precision required for identifier grounding tasks. Models must produce exact matches to succeed, as approximate or similar identifiers are incorrect. This task represents the most challenging capability in MetaBench, with even the best-performing models (\textit{Claude-sonnet-4}) achieving only 0.87\% accuracy without retrieval augmentation, highlighting a fundamental limitation in current LLM architectures for structured database mapping tasks.

\section{Prompt for reasoning benchmark}
\label{sec:appendix-nl-generation}
To construct the Reasoning (Triple Extraction) benchmark, we generated natural language sentences from structured knowledge graph triples using DeepSeek-V3.1~\citep{deepseekv3}. Each triple from MetaKG was transformed into a natural language sentence using the prompt template shown below. The prompt explicitly instructs the model to avoid template-like language and to create varied, natural expressions of biomedical facts, ensuring that the resulting benchmark requires genuine semantic parsing rather than pattern matching.
\begin{tcolorbox}[
title=\textbf{Prompt for Reasoning Benchmark},
colframe=black!50,
colback=black!5,
boxrule=0.5pt,
arc=2pt,
fonttitle=\bfseries,
breakable
]
Given the following triple from a biomedical knowledge graph:\\
\textbf{Subject}: {head}\\
\textbf{Predicate}: {rel}\\
\textbf{Object}: {tail}\\
Write a natural language sentence that expresses this fact, but do so in a way that is different from a simple template. Be creative and vary the phrasing. Do not use the words 'subject', 'predicate', or 'object'.
\end{tcolorbox}
This generation approach ensures that the Reasoning task evaluates genuine natural language understanding rather than template recognition. The diversity in expression styles reflects the variability found in real scientific literature, making the benchmark more representative of practical knowledge extraction scenarios in metabolomics research.

\section{Triple extraction examples}
\label{sec:appendix-reasoning-examples}
The Reasoning capability is assessed through knowledge graph triple extraction tasks derived from MetaKG. Models receive natural language sentences describing metabolite relationships and must extract structured triples in the format (head, relationship, tail). This task evaluates the model's ability to parse unstructured scientific text and perform structured reasoning, which is a critical skill for automated knowledge extraction from metabolomics literature. Below are representative examples demonstrating disease associations and metabolic disorder relationships.
\begin{tcolorbox}[
title=\textbf{Example 1: Disease Association},
colframe=black!50,
colback=black!5,
boxrule=0.5pt,
arc=2pt,
fonttitle=\bfseries,
breakable
]
\textbf{Text:} Research has linked L-Methionine to the onset and manifestation of Epilepsy.

\textbf{Extracted Triple:}
\begin{itemize}
\item \textbf{Head:} L-Methionine
\item \textbf{Relationship:} has\_disease
\item \textbf{Tail:} Epilepsy
\end{itemize}
\end{tcolorbox}
\begin{tcolorbox}[
title=\textbf{Example 2: Metabolic Disorder},
colframe=black!50,
colback=black!5,
boxrule=0.5pt,
arc=2pt,
fonttitle=\bfseries,
breakable
]
\textbf{Text:} Capric acid is implicated in the development of disorders related to metabolism and nutrition.

\textbf{Extracted Triple:}
\begin{itemize}
\item \textbf{Head:} Capric acid
\item \textbf{Relationship:} has\_disease
\item \textbf{Tail:} Metabolism and nutrition disorders
\end{itemize}
\end{tcolorbox}
\begin{tcolorbox}[
title=\textbf{Example 3: Biomarker Relationship},
colframe=black!50,
colback=black!5,
boxrule=0.5pt,
arc=2pt,
fonttitle=\bfseries,
breakable
]
\textbf{Text:} Citrulline serves as a crucial biomarker for the diagnosis and assessment of intestinal failure.

\textbf{Extracted Triple:}
\begin{itemize}
\item \textbf{Head:} Citrulline
\item \textbf{Relationship:} has\_disease
\item \textbf{Tail:} Intestinal failure
\end{itemize}
\end{tcolorbox}
\begin{tcolorbox}[
title=\textbf{Example 4: Deficiency Manifestation},
colframe=black!50,
colback=black!5,
boxrule=0.5pt,
arc=2pt,
fonttitle=\bfseries,
breakable
]
\textbf{Text:} A deficiency in Vitamin A can manifest as a range of afflictions impacting the nervous system.

\textbf{Extracted Triple:}
\begin{itemize}
\item \textbf{Head:} Vitamin A
\item \textbf{Relationship:} has\_disease
\item \textbf{Tail:} Nervous system disorders
\end{itemize}
\end{tcolorbox}
\begin{tcolorbox}[
title=\textbf{Example 5: Contributing Factor},
colframe=black!50,
colback=black!5,
boxrule=0.5pt,
arc=2pt,
fonttitle=\bfseries,
breakable
]
\textbf{Text:} A deficiency in propionylcarnitine can be a contributing factor in the development of missing teeth.

\textbf{Extracted Triple:}
\begin{itemize}
\item \textbf{Head:} Propionylcarnitine
\item \textbf{Relationship:} has\_disease
\item \textbf{Tail:} Missing teeth
\end{itemize}
\end{tcolorbox}
\vspace{0.5cm}

\section{Study description generation) examples}
\label{sec:appendix-research-examples}
The Research capability is assessed through study description generation tasks derived from MetaboLights. Models receive concise study titles as input and must generate comprehensive descriptions containing experimental methodologies, analytical techniques, key findings, and biological implications. This task represents the highest level of scientific complexity in MetaBench, testing the model's ability to synthesize knowledge about research design, technical protocols, and scientific interpretation. Below are representative examples demonstrating diverse research domains in metabolomics.
\begin{tcolorbox}[
title=\textbf{Example 1: Software Evaluation},
colframe=black!50,
colback=black!5,
boxrule=0.5pt,
arc=2pt,
fonttitle=\bfseries,
breakable
]
\textbf{Question:} Write a study description for the study titled 'Comprehensive evaluation of untargeted metabolomics data processing software in feature detection, quantification and discriminating marker selection'.

\textbf{Reference Description (excerpt):}
Data analysis represents a key challenge for untargeted metabolomics studies and it commonly requires extensive processing of more than thousands of metabolite peaks included in raw high-resolution MS data. Although a number of software packages have been developed to facilitate untargeted data processing, they have not been comprehensively scrutinized in the capability of feature detection, quantification and marker selection using a well-defined benchmark sample set. In this study, we acquired a benchmark dataset from standard mixtures consisting of 1100 compounds with specified concentration ratios including 130 compounds with significant variation of concentrations. Five software evaluated here (MS-Dial, MZmine 2, XCMS, MarkerView, and Compound Discoverer) showed similar performance in detection of true features... [continues]
\end{tcolorbox}
\begin{tcolorbox}[
title=\textbf{Example 2: Clinical Metabolomics},
colframe=black!50,
colback=black!5,
boxrule=0.5pt,
arc=2pt,
fonttitle=\bfseries,
breakable
]
\textbf{Question:} Write a study description for the study titled 'Altered metabolome and microbiome features provide clues in understanding irritable bowel syndrome and depression comorbidity'.

\textbf{Reference Description (excerpt):}
Irritable bowel syndrome (IBS) is one of the functional gastrointestinal disorders characterized by chronic and/or recurrent symptoms of abdominal pain and irregular defecation. Changed gut microbiota has been proposed to mediate IBS; however, contradictory results exist, and IBS-specific microbiota, metabolites, and their interactions remain poorly understood. To address this issue, we performed metabolomic and metagenomic profiling of stool and serum samples based on discovery (n = 330) and validation (n = 101) cohorts... [continues]
\end{tcolorbox}
\begin{tcolorbox}[
title=\textbf{Example 3: Multi-Omics Integration},
colframe=black!50,
colback=black!5,
boxrule=0.5pt,
arc=2pt,
fonttitle=\bfseries,
breakable
]
\textbf{Question:} Write a study description for the study titled 'Mechanisms of hepatic steatosis in chickens: integrated analysis of the host genome, molecular phenomes and gut microbiome'.

\textbf{Reference Description (excerpt):}
Hepatic steatosis is the initial manifestation of abnormal liver functions and often leads to liver diseases such as non-alcoholic fatty liver disease in humans and fatty liver syndrome in animals. In this study, we conducted a comprehensive analysis of a large chicken population consisting of 705 adult hens by combining host genome resequencing, liver transcriptome, proteome and metabolome analysis, as well as microbial 16S rRNA gene sequencing of each gut segment... [continues]
\end{tcolorbox}
\begin{tcolorbox}[
title=\textbf{Example 4: Developmental Metabolomics},
colframe=black!50,
colback=black!5,
boxrule=0.5pt,
arc=2pt,
fonttitle=\bfseries,
breakable
]
\textbf{Question:} Write a study description for the study titled 'Changes in the Milk Metabolome of the Giant Panda (Ailuropoda melanoleuca) with Time after Birth – Three Phases in Early Lactation and Progressive Individual Differences'.

\textbf{Reference Description (excerpt):}
Ursids (bears) in general, and giant pandas in particular, are highly altricial at birth. The components of bear milks and their changes with time may be uniquely adapted to nourish relatively immature neonates, protect them from pathogens, and support the maturation of neonatal digestive physiology. Serial milk samples collected from three giant pandas in early lactation were subjected to untargeted metabolite profiling and multivariate analysis... [continues]
\end{tcolorbox}
\begin{tcolorbox}[
title=\textbf{Example 5: Subcellular Imaging},
colframe=black!50,
colback=black!5,
boxrule=0.5pt,
arc=2pt,
fonttitle=\bfseries,
breakable
]
\textbf{Question:} Write a study description for the study titled 'Subcellular antibiotic visualization reveals a dynamic drug reservoir in infected macrophages'.

\textbf{Reference Description (excerpt):}
Tuberculosis, caused by the intracellular pathogen Mycobacterium tuberculosis, remains the world's deadliest infectious disease. Sterilizing chemotherapy requires at least 6 months of multidrug therapy. Difficulty visualizing the subcellular localization of antibiotics in infected host cells means that it is unclear whether antibiotics penetrate all mycobacteria-containing compartments in the cell. Here, we combined correlated light, electron, and ion microscopy to image the distribution of bedaquiline in infected human macrophages... [continues]
\end{tcolorbox}
\vspace{0.5cm}

\section{Compared LLMs}
\label{app:llms}
\begin{table*}[t]
\centering
\begin{threeparttable}
\caption{Models evaluated in MetaBench. Parameter counts are from official sources where available; proprietary models do not disclose parameter counts.}
\label{tab:models_metabench}
\setlength{\tabcolsep}{7pt}
\renewcommand{\arraystretch}{0.75}
\setlength{\heavyrulewidth}{0.9pt}
\setlength{\lightrulewidth}{0.4pt}
\begin{tabular}{@{}lllll@{}}
\toprule
\rowcolor{headerblue}
\textbf{Model} & \textbf{Params} & \textbf{Architecture} & \textbf{Release date} & \textbf{Provider} \\
\midrule
\rowcolor{black!6}\textbf{Open-source models} & & & & \\
Qwen3-1.7b & 1.7B & dense & 2025-04-29 & Alibaba Qwen \\
Qwen3-4b   & 4B   & dense & 2025-04-29 & Alibaba Qwen \\
Qwen3-8b   & 8B   & dense & 2025-04-29 & Alibaba Qwen \\
Qwen3-14b  & 14B  & dense & 2025-04-29 & Alibaba Qwen \\
Qwen3-32b  & 32B  & dense & 2025-04-29 & Alibaba Qwen \\
\midrule
Gemma-3-270m-it & 270M & dense & 2025-03-12 & Google \\
Gemma-3-1b-it   & 1B   & dense & 2025-03-12 & Google \\
Gemma-3-4b-it   & 4B   & dense & 2025-03-12 & Google \\
Gemma-3-12b     & 12B  & dense & 2025-03-12 & Google \\
Gemma-3-27b     & 27B  & dense & 2025-03-12 & Google \\
\midrule
Llama-3.2-1b    & 1B   & dense & 2024-09-25 & Meta \\
Llama-3.2-3b    & 3B   & dense & 2024-09-25 & Meta \\
Llama-3.1-8b    & 8B   & dense & 2024-07-23 & Meta \\
Llama-3.1-70b   & 70B  & dense & 2024-07-23 & Meta \\
\midrule
DeepSeek-v3.1   & 685B (MoE)\tnote{a} & MoE & 2025-08-21 & DeepSeek \\
\midrule
GPT-oss-20b   & 20B & MoE & 2025-08-21 & OpenAI \\
\midrule
\rowcolor{black!6}\textbf{Closed-source models} & & & & \\
GPT-4o-mini     & Not disclosed & multimodal & 2024-07-18 & OpenAI \\
GPT-5-nano      & Not disclosed & unified/thinking variants & 2025-08-07 & OpenAI \\
GPT-5-mini      & Not disclosed & unified/thinking variants & 2025-08-07 & OpenAI \\
GPT-5           & Not disclosed & unified/thinking variants & 2025-08-07 & OpenAI \\
\midrule
Claude-haiku-3.5  & Not disclosed & hybrid family     & 2024-10-22 & Anthropic \\
Claude-sonnet-3.7 & Not disclosed & hybrid reasoning  & 2025-02-24 & Anthropic \\
Claude-sonnet-4   & Not disclosed & hybrid reasoning  & 2025-05-22 & Anthropic \\
\midrule
Gemini-2.5-flash & Not disclosed & thinking model & 2025-06-17 & Google \\
Gemini-2.5-pro   & Not disclosed & thinking model & 2025-03-25 & Google \\
\bottomrule
\end{tabular}
\begin{tablenotes}[flushleft]
\footnotesize
\item[a] MoE = mixture-of-experts. Count refers to total parameters across experts, not active per token.
\end{tablenotes}
\end{threeparttable}
\end{table*}

We evaluate 25 state-of-the-art large language models spanning eight model families to ensure comprehensive coverage of current LLM capabilities in metabolomics. Our selection includes both open-source models, \textit{Qwen3} (1.7B–32B), \textit{Gemma-3} (270M–27B), \textit{Llama} (1B–70B), \textit{DeepSeek-v3.1} (685B MoE), and \textit{GPT-oss-20b} (20B MoE)and closed-source systems from leading providers: OpenAI's \textit{GPT} series (4o-mini through GPT-5), Anthropic's \textit{Claude} family (haiku-3.5, sonnet-3.7, sonnet-4), and Google's \textit{Gemini-2.5} models (flash and pro). This diversity enables systematic analysis of how model scale, architecture (dense vs. mixture-of-experts), and training paradigms affect performance across the five metabolomics capabilities assessed in MetaBench. \textbf{Table}~\ref{tab:models_metabench} summarizes the key specifications of all evaluated models, including parameter counts (where available), architectural types, release dates, and providers. This comprehensive model selection establishes MetaBench as a rigorous testbed for tracking progress in metabolomics-oriented language modeling across both academic and industrial research efforts.

\section{System prompts}
\label{app:system_prompts}

In our evaluation framework, we designed five distinct system prompts, each tailored to a capability assessed in MetaBench.  
These prompts ensure that large language models receive clear, domain-specific instructions aligned with metabolomics tasks.  

\begin{itemize}
  \item \textbf{MCQA (Multiple-Choice Question Answering):} Tests factual knowledge of metabolites, pathways, enzymes, and biological systems.  
  \item \textbf{Triple Extraction:} Assesses reasoning ability by extracting \textit{(head, relationship, tail)} triples from natural language.  
  \item \textbf{Study Description:} Evaluates scientific writing and comprehension by generating detailed study descriptions from research titles.  
  \item \textbf{QA (Open-ended Question Answering):} Probes explanatory ability on pathways, mechanisms, and processes.  
  \item \textbf{Metabolite Mapping:} Measures precision in identifier mapping across metabolomics databases like KEGG, HMDB and ChEBI.  
\end{itemize}

The following parts contain the exact system prompts used for each evaluation task.

\begin{tcolorbox}[
    title=\textbf{System Prompt for MCQA},
    colframe=black!50,
    colback=black!5,
    boxrule=0.5pt,
    arc=2pt,
    fonttitle=\bfseries,
    breakable
]

\textbf{Task:}  

Answer multiple choice questions about metabolite properties, pathways, enzymes, and biological processes.  

\textbf{Instructions:}  

- You will be given a question and four options (A, B, C, D).  \\
- Choose the correct answer based on your knowledge of metabolomics, biochemistry, and biological systems.  \\
- Respond with only the letter (A, B, C, or D).  

\textbf{Coverage:}  

- Metabolite taxonomy and classification \\ 
- Enzyme associations and EC numbers  \\
- KEGG pathway and network associations  \\
- Molecular properties (weight, structure) \\ 
- Tissue locations and biological functions \\ 
- Protein and gene relationships  

\textbf{Examples:}  

Q: What is the taxonomy class of Prostaglandin E1?  

Options: \{A: Prostaglandins and related compounds, B: Phenylpyrazoles, C: Furanones, D: Azolines\}  \\
Answer: A  

----------------------------------------------------

Q: What is the taxonomy class of PC(18:1(9Z)/18:0)? 

Options: \{A: Alanine and derivatives, B: Glycerophospholipids, C: Anisoles, D: Isoleucine and derivatives\} \\ 
Answer: B  

----------------------------------------------------

Q: Where is Acetylcholine located in tissue?  
Options: \{A: Bone Marrow, B: Adrenal Cortex, C: Vitreous humor, D: Neuron\}  \\
Answer: D  

\end{tcolorbox}

\begin{tcolorbox}[title=\textbf{System Prompt for Triple Extraction},     colframe=black!50,
    colback=black!5,
    boxrule=0.5pt,
    arc=2pt,
    fonttitle=\bfseries,
    breakable]

\textbf{Task:}  

Extract knowledge graph triples from natural language text about metabolites and their biological properties.  

\textbf{Format:}  

Head: [metabolite/entity]  \\
Relationship: [relationship type]  \\
Tail: [disease/condition/property/location]  

\textbf{Allowed relationships:}  

- \verb|has\_disease| : metabolite associated with diseases/conditions  \\
- \verb|has\_disposition| : metabolite location or cellular disposition  \\
- \verb|has\_smiles| : molecular structure in SMILES notation  \\
- \verb|has\_synonym| : alternative names  \\
- \verb|has\_class| : biochemical classification  \\
- \verb|has\_tissue_location| : tissue or organ presence  

\textbf{Examples:}  

Text: Within the human body, the blood's platelets are a known reservoir for the compound inosine.  \\
Head: Inosine  \\
Relationship: has\_tissue\_location  \\
Tail: Platelet  

----------------------------------------------------

Text: Dolichyl phosphate D-mannose is categorized among the broader class of carbohydrates and carbohydrate conjugates. \\ 
Head: Dolichyl phosphate D-mannose  \\
Relationship: has\_class  \\
Tail: Carbohydrates and carbohydrate conjugates  

----------------------------------------------------

Text: The molecular structure of Isobutyryl-L-carnitine is uniquely defined by the SMILES string \verb|CC(C)C(=O)O[C@H](CC(O)=O)...|  \\
Head: Isobutyryl-L-carnitine \\
Relationship: has\_smiles  \\
Tail: CC(C)C(=O)O[C@H](CC(O)=O)...

\end{tcolorbox}

\begin{tcolorbox}[title=\textbf{System Prompt for Study Description},     colframe=black!50,
    colback=black!5,
    boxrule=0.5pt,
    arc=2pt,
    fonttitle=\bfseries,
    breakable]

\textbf{Task:}  

Write a comprehensive study description for a given metabolomics research title.  

\textbf{Instructions:}  

Include the following:  \\
1. Study purpose and objectives  \\
2. Methodology and experimental design (LC-MS, GC-MS, NMR)  \\
3. Key findings and results  \\
4. Clinical or scientific significance  \\
5. Sample characteristics and analytical methods  \\
6. Metabolic pathways and biological processes involved  

\textbf{Coverage:}  

- Untargeted and targeted metabolomics  \\
- Sample preparation and analytical techniques  \\
- Data processing and statistical analysis  \\
- Biological interpretation and pathway analysis  \\
- Clinical applications and biomarker discovery  

\textbf{Example:}  

Title: "Dissolved organic carbon compounds in deep-sea hydrothermal vent fluids from the East Pacific Rise at ...  \\
Description: "... detailed description ..."  

\end{tcolorbox}

\begin{tcolorbox}[title=\textbf{System Prompt for QA},     colframe=black!50,
    colback=black!5,
    boxrule=0.5pt,
    arc=2pt,
    fonttitle=\bfseries,
    breakable]

\textbf{Task:}  

Answer questions about biochemical definitions, metabolic pathways, and biological processes.  

\textbf{Instructions:}  

Provide comprehensive, accurate answers including:  \\
1. Definition of the pathway, process, or concept  \\
2. Key enzymes, reactions, and components \\ 
3. Biological significance and function  \\
4. Clinical or pathological relevance (if any)  \\
5. Related pathways or mechanisms  

\textbf{Coverage:}  

- Metabolic pathways (TCA cycle, glycolysis, etc.)  \\
- Biochemical processes and mechanisms  \\
- Disease definitions and mechanisms  \\
- Drug action pathways  \\
- Operon regulation  \\
- Molecular signaling  

\textbf{Examples:}  

Q: What is the definition of Succinate Signalling?  

A: "... detailed definition ..."  

----------------------------------------------------

Q: What is the definition of Tetracycline Action Pathway?  

A: "... detailed definition ..."  

\end{tcolorbox}

\begin{tcolorbox}[title=\textbf{System Prompt for Metabolite Mapping},     colframe=black!50,
    colback=black!5,
    boxrule=0.5pt,
    arc=2pt,
    fonttitle=\bfseries,
    breakable]

\textbf{Task:}  

Answer metabolite identifier mapping questions between databases and naming conventions.  

\textbf{Instructions:}  

- Provide only the identifier value as the answer.  \\
- No extra explanation or formatting.  

\textbf{Supported identifiers:} 

- CAS Registry Numbers  \\
- PubChem CIDs  \\
- Chemical names (IUPAC, common) \\ 
- HMDB IDs  \\
- ChEBI IDs  \\
- KEGG IDs  \\

\textbf{Examples:}  

Q: What is the KEGG ID of metabolite Tamibarotene?  

A: \verb|C12864|  

----------------------------------------------------

Q: What is the HMDB ID of metabolite TG(15:0/i-20:0/a-17:0)[rac]?  

A: \verb|HMDB0102782|  

----------------------------------------------------

Q: What is the ChEBI ID of Benzilic acid?

A: \verb|39414|  

----------------------------------------------------

Q: What is the KEGG ID of 2-Iodophenol?  

A: \verb|C01874|  

\end{tcolorbox}

\section{The Critical Role of Metabolite Identifier Grounding in Metabolomics}
\label{sec:appendix-grounding-importance}
Metabolite identifier groundingt represents a fundamental bottleneck in metabolomics research~\citep{swainston2014comparative}. A single metabolite may be referenced by multiple identifiers: KEGG uses compound IDs (e.g., C00031 for D-Glucose), HMDB employs alphanumeric codes (e.g., HMDB0000122), ChEBI assigns numerical IDs (e.g., 17234), and scientific literature uses IUPAC names, common names, or synonyms, with PubChem listing nearly 250 synonyms for L-Tryptophan alone~\citep{krettler2024navigating}. This heterogeneity arose because different databases emerged from distinct research communities: KEGG emphasizes pathway context, HMDB focuses on human metabolites with clinical relevance, ChEBI provides chemical ontology, and PubChem offers comprehensive chemical data~\citep{wishart2022hmdb, kanehisa2025kegg}. Each adopted identifier schemes suited to their organization, predating standardization efforts.

Accurate identifier grounding is essential throughout the research pipeline. Cross-study meta-analyses depend on harmonizing identifiers across laboratories and platforms~\citep{braisted2023metlinkr}. Database construction efforts like MetaKG require linking millions of facts from heterogeneous sources to unified metabolite entities~\citep{lu2025knowledge}. Identifier mapping errors cascade through research quality: incorrect links lead to spurious discoveries, failed recognition causes missed connections, and inconsistent usage prevents reproducible science~\citep{cajka2022reproducibility, powers2024checklist}.

Database inconsistency compounds these challenges. Analysis of 11 biochemical databases revealed that while HMDB maintains high consistency (only 1.7\% of names linked to multiple IDs), ChEBI and KEGG show 14.8\% and 13.3\% ambiguity respectively~\citep{grapov2018consistency}. Inter-database mapping using metabolite names shows inconsistencies ranging from 0\% to 81.2\%, with similar results (0-83\%) when mapping via reference identifiers~\citep{grapov2018consistency}. Current solutions include manually curated mapping tables (e.g., MetabolitesID~\citep{metaboliteidmapping}), web-based conversion tools like the Chemical Translation Service~\citep{wohlgemuth2010chemical}, and APIs, but these face persistent challenges: no single resource provides comprehensive coverage, databases update frequently requiring continuous curation, automated tools struggle with isomers and context-dependent synonyms, and different tools employ incompatible interfaces~\citep{swainston2014comparative, krettler2024navigating}. Large Language Models promise flexible, context-aware resolution leveraging both structured databases and unstructured literature. However, MetaBench demonstrates that current models exhibit catastrophic failure without retrieval augmentation, achieving less than 1\% accuracy even for frontier architectures. This finding establishes identifier grounding as a critical bottleneck impeding the computational transformation of metabolomics toward integrated, AI-assisted discovery systems.

\end{document}